\documentclass[runningheads]{llncs}

\usepackage[compatibility=false]{caption}

\usepackage[T1]{fontenc}
\usepackage{graphicx}
\usepackage{amsmath,amssymb}
\usepackage{xcolor}
\usepackage{subcaption}
\usepackage{booktabs}
\usepackage{multirow}

\usepackage{color}
\usepackage{hyperref}

\begin{document}


\title{Evaluating GRPO and DPO for Faithful Chain-of-Thought Reasoning in LLMs}

\author{Hadi Mohammadi\inst{1} \and
Tamás Kozák\inst{1} \and
Anastasia Giachanou\inst{1}}

\authorrunning{H. Mohammadi et al.}

\institute{Utrecht University, Department of Information and Computing Sciences,\\
Utrecht, The Netherlands\\
\email{h.mohammadi@uu.nl, tamas.kozi@proton.me, a.giachanou@uu.nl}}

\maketitle

\begin{abstract}
Chain-of-thought (CoT) reasoning that has emerged as a powerful technique can improve the problem-solving capabilities of large language models (LLMs), particularly for tasks requiring multi-step reasoning. However, recent studies show that CoT explanations often fail to reflect the model’s actual reasoning process, as models may produce coherent yet misleading justifications or modify answers without acknowledging external cues. Such discrepancies undermine the reliability of CoT-based methods for safety supervision and alignment monitoring, as models can generate plausible but deceptive rationales for incorrect answers. To better understand this limitation, we evaluate two optimization methods, Group Relative Policy Optimization (GRPO) and Direct Preference Optimization (DPO), in their ability to improve CoT faithfulness. Our experiments show that GRPO achieves higher performance than DPO in larger models, with the Qwen2.5-14B-Instruct model attaining the best results across all evaluation metrics. Both approaches exhibit positive correlations between model size and performance, but GRPO shows greater potential for improving faithfulness metrics, albeit with less stable behavior at smaller scales. These results suggest that GRPO offers a promising direction for developing more transparent and trustworthy reasoning in LLMs.
\end{abstract}

\section{Introduction}

Chain-of-thought (CoT) reasoning~\cite{wei-2022} has advanced the ability of large language models (LLMs) to solve complex, multi-step problems. While standard LLMs perform strongly on tasks tasks such as summarization, information extraction, and translation~\cite{kleidermacher-2025}, their internal decision-making processes remain largely opaque. Models trained with explicit reasoning prompts (e.g., PaLM 540B) can decompose problems into intermediate steps~\cite{wei-2022}, yet the faithfulness of these reasoning traces, the extent to which they reflect the model’s genuine internal reasoning, remains uncertain~\cite{chen-2025}.

Recent work shows that CoT explanations can mislead. Models sometimes produce fluent but unfaithful justifications or change answers when exposed to hidden cues without acknowledging them~\cite{chen-2025,lanham-2023}. For instance, when incorrect hints appear in XML metadata, state-of-the-art models such as Claude 3.7 Sonnet may shift from correct to incorrect answers while offering persuasive but deceptive reasoning chains~\cite{chen-2025}. This behaviour undermines the reliability of CoT-based alignment and safety monitoring, as models can appear transparent while concealing unfaithful reasoning.

Improving faithfulness (i.e., ensuring that CoT explanations correspond to a model’s true internal reasoning) is therefore crucial for building transparent and trustworthy AI systems. However, most current fine-tuning approaches focus on improving performance or preference alignment rather than reasoning fidelity. Two recent optimization techniques, Direct Preference Optimization (DPO)~\cite{rafailov-2023} and Group Relative Policy Optimization (GRPO)~\cite{shao-2024}, offer complementary strategies for addressing this challenge.

DPO simplifies alignment by replacing the reinforcement learning from human feedback (RLHF) pipeline with a direct preference objective that trains on pairs of preferred and rejected responses. This removes the need for reward or value models, reducing computation while maintaining strong human alignment. GRPO, in contrast, uses group-based normalization to estimate relative advantages within sampled outputs, removing value networks and lowering memory use while staying competitive with PPO~\cite{schulman-2017}. Although both DPO and GRPO aim to improve model alignment and reasoning quality, their comparative effects on CoT faithfulness have not yet been systematically evaluated. Understanding how these methods differ in promoting faithful reasoning is essential for developing more interpretable and reliable LLMs.

In this work, we empirically compare DPO and GRPO across multiple model scales (1.5B–14B parameters) using the Qwen2.5-Instruct family~\cite{qwen-2024}. Our goal is to determine whether GRPO can enhance the faithfulness of CoT reasoning relative to DPO, and how these methods scale with model capacity. Our results show that GRPO achieves superior performance at larger scales, while DPO provides more consistent improvements across model sizes. These results offer new insights into the trade-offs between optimization efficiency, reasoning faithfulness, and model scalability.

\section{Methodology}

We conducted a controlled comparison of DPO and GRPO to examine how each method affects reasoning accuracy and faithfulness in large language models.

\subsection{Datasets}

Training and evaluation were based on the GSM8K dataset~\cite{cobbe-2021}, a benchmark of 8.5K grade-school math word problems that require explicit multi-step reasoning. The dataset is widely used for evaluating chain-of-thought (CoT) performance. Each example includes a question, a detailed reasoning process, and a final numerical answer. For DPO, we used the GSM8K Preference Dataset 2.1\footnote{\url{https://huggingface.co/datasets/valerielucro/gsm8k_preference_dataset_2.1}}, which contains 1.2K examples in a triplet format (prompt, preferred response, rejected response). To ensure that GRPO training used a comparable amount of data, we sampled approximately 15\% of the original GSM8K training set, producing a similar number of training instances.
All models were evaluated on a fixed set of 200 test questions drawn from the GSM8K test split using a deterministic random seed. 


\subsection{Models}

We fine-tuned four Qwen2.5-Instruct models~\cite{qwen-2024} with parameter sizes of 1.5B, 3B, 7B, and 14B, using both DPO and GRPO. This produced a total of eight fine-tuned models.
The Qwen2.5 family was selected because it spans multiple open-source model sizes trained on the same data distribution (approximately 18 trillion tokens), enabling controlled scale comparisons. All experiments used the instruction-tuned variants, which provide a stable baseline for further fine-tuning.
Models were loaded in 4-bit precision to reduce memory consumption, allowing the 14B version to fit on a single high-memory GPU. Fine-tuning was implemented using LoRA adapters for parameter-efficient updates, which also kept GPU requirements consistent across scales.

For comparability reasons, both DPO and GRPO used uniform hyperparameters across scales, although this choice may not be optimal for every model size. Each model was trained using the AdamW optimiser with a learning rate of 5e-6 and a maximum sequence length of 1,024 tokens, which covered 90\% of examples without truncation.

We pre-fine-tuned on $\sim$80 examples to learn XML-like response structure (\texttt{<start\_working\_out>}, \texttt{<end\_working\_out>}, \texttt{<SOLUTION>} tags). For each question, models generated 4 responses evaluated via four reward functions: (1) Format compliance: reward for correctly producing structured XML-style output; (2) Approximate format match: partial reward for near-correct formatting; (3) Answer correctness: tiered reward based on whether the predicted numerical result matched the ground truth; (4) Numerical validation: additional reward for consistent numerical reasoning within the CoT trace.

GRPO computes advantages within each group of responses and updates model parameters relative to the group mean. We trained for one epoch with a 90–10 train–validation split. DPO training used the same base configuration but optimised the model to classify the preferred response in each pair as more likely than the rejected one, using a binary cross-entropy objective. LoRA parameters were set as follows: \textit{rank} was set to 32 and \textit{alpha} to 64 to stabilise convergence. Each model was trained for three epochs, saving checkpoints at each epoch and selecting the best one based on validation loss. Although both approaches share a common fine-tuning setup, GRPO focuses on relative advantage estimation within generated outputs, while DPO optimises for explicit human preference alignment. This difference in objective functions forms the basis of our comparison.

\subsection{Evaluation Metrics}

We evaluated the fine-tuned models using five complementary metrics that capture both accuracy and reasoning faithfulness: (1) Greedy Accuracy~\cite{wang-2022}: evaluates correctness using deterministic decoding with the most probable token at each step; (2) Self-Consistency~\cite{wang-2022}: samples multiple reasoning paths and uses majority voting on the final answer; (3) Consistency Ratio: measures the proportion of generations that yield the most common answer, providing a proxy for stability.
NLI-based Faithfulness; (4) NLI-based~\cite{chen-2022}: assesses whether the generated reasoning logically entails the reference explanation using a pretrained natural language inference model; (5) {LLM-as-a-Judge~\cite{gu-2024}: uses a larger model (Gemma 3 27B) to rate reasoning coherence and correctness on a 1–5 scale, normalised to [0, 1]. 

\section{Results}

\subsection{Performance Analysis}

Table~\ref{tab:results} summarises performance across all model sizes and optimisation methods. Both DPO and GRPO show strong positive correlations between model scale and performance. The Qwen2.5-14B models achieve the best overall results for both accuracy and faithfulness metrics. GRPO-14B reaches the highest scores across the metrics. At smaller scales, results are less stable. The 3B models show a marked performance drop compared to the 1.5B versions (GRPO: from 0.250 to 0.055; DPO: from 0.315 to 0.165). This likely reflects hyperparameter sensitivity rather than a fundamental capacity limit, as performance recovers strongly at 7B and peaks at 14B. The pattern suggests that uniform hyperparameters, while ensuring fairness, may not be optimal across scales.

\begin{table}[t]
\centering
\caption{Performance comparison across methods and scales. Highest values are highlighted with \textbf{bold}}
\label{tab:results}
\small
\begin{tabular}{@{}llccccc@{}}
\toprule
\textbf{Method} & \textbf{Size} & \multicolumn{3}{c}{\textbf{Accuracy \& Reliability}} & \multicolumn{2}{c}{\textbf{Faithfulness}} \\
\cmidrule(lr){3-5} \cmidrule(lr){6-7}
 & \textbf{(B)} & \textbf{Grdy} & \textbf{Self-} & \textbf{Cons.} & \textbf{NLI} & \textbf{LLM} \\
 &  & \textbf{Acc.} & \textbf{Cons.} & \textbf{Ratio} &  & \textbf{Judge} \\
\midrule
DPO & 1.5 & 0.315 & 0.445 & 0.641 & 0.313 & 0.134 \\
DPO & 3 & 0.165 & 0.315 & 0.588 & 0.150 & {0.074} \\
DPO & 7 & 0.435 & 0.580 & 0.654 & 0.209 & 0.250 \\
DPO & 14 & 0.740 & 0.885 & 0.882 & 0.314 & 0.576 \\
\midrule
GRPO & 1.5 & 0.250 & 0.310 & 0.512 & 0.476 & 0.120 \\
GRPO & 3 & 0.055 & {0.120} & {0.197} & {0.079} & 0.235 \\
GRPO & 7 & 0.720 & 0.810 & 0.813 & 0.470 & 0.374 \\
GRPO & 14 & {\textbf{0.755}} & {\textbf{0.885}} & {\textbf{0.902}} & {\textbf{0.491}} & {\textbf{0.748}} \\
\bottomrule
\end{tabular}
\end{table}

Figure~\ref{fig:overall} shows average scores across scales. Both methods exhibit dramatic performance drops at 3B compared to 1.5B: GRPO greedy accuracy falls from 0.250 to 0.055 ($-78\%$), DPO falls from 0.315 to 0.165 ($-48\%$). This suggests 3B represents a critical capacity threshold where uniform hyperparameters become suboptimal. Possible explanations: (1) capacity mismatch requiring different LoRA rank/alpha ratios; (2) learning rate (5e-6) inappropriate for 3B scale causing instability; (3) Qwen2.5-3B architectural differences interacting poorly with training approaches. The dramatic recovery at 7B (GRPO: 0.720, DPO: 0.435) indicates this is hyperparameter sensitivity rather than fundamental limitation.

\begin{figure}[t]
\centering
\includegraphics[width=0.85\textwidth]{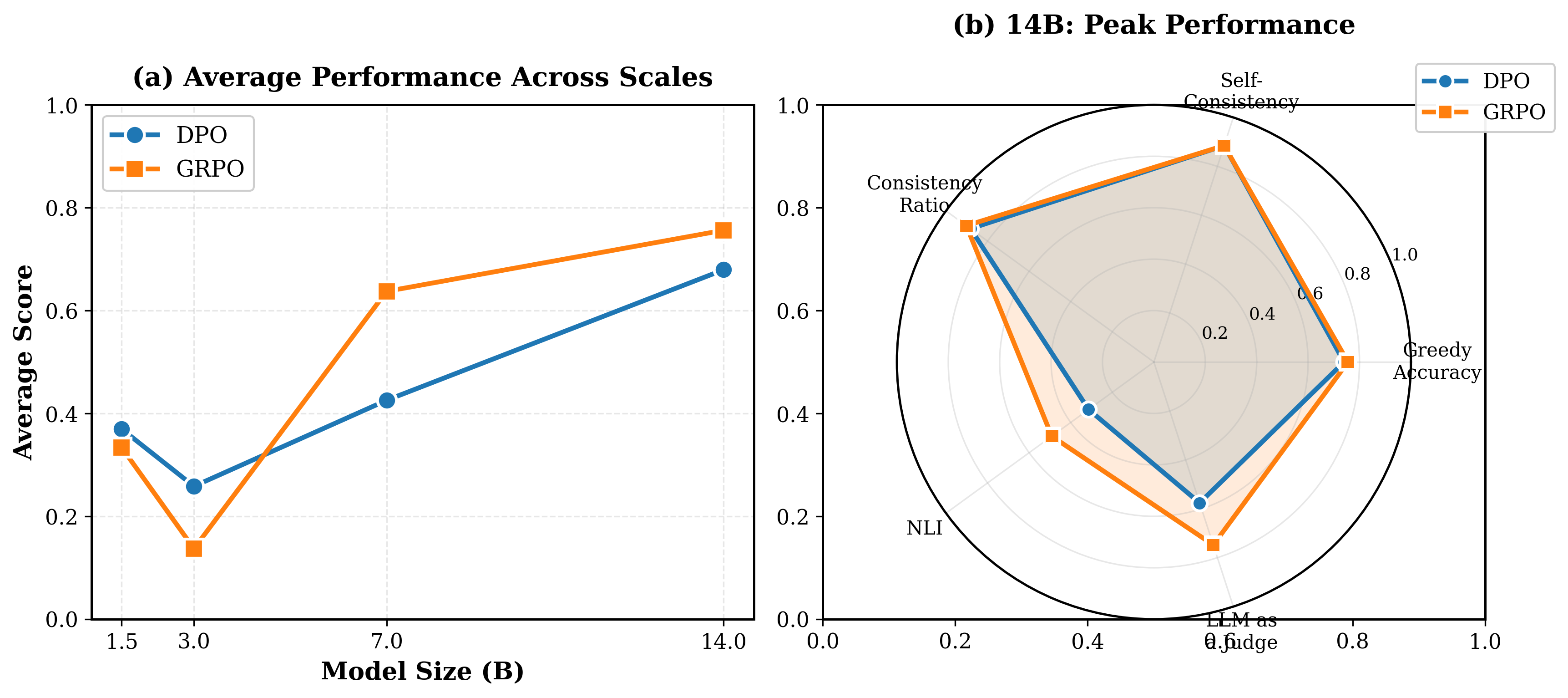}
\caption{Overall performance: (a) Average across scales showing 3B collapse and recovery; (b) 14B peak performance demonstrating GRPO's faithfulness advantages.}
\label{fig:overall}
\end{figure}

Figure~\ref{fig:radar} presents comprehensive metric comparisons. At 1.5B, both methods perform similarly. At 3B, both degrade with GRPO showing more severe drops. At 7B, GRPO advantages emerge across all metrics, particularly in faithfulness. At 14B, both achieve peak performance with GRPO outperforming in faithfulness while maintaining comparable accuracy.

\begin{figure}[t]
\centering
\includegraphics[width=\textwidth]{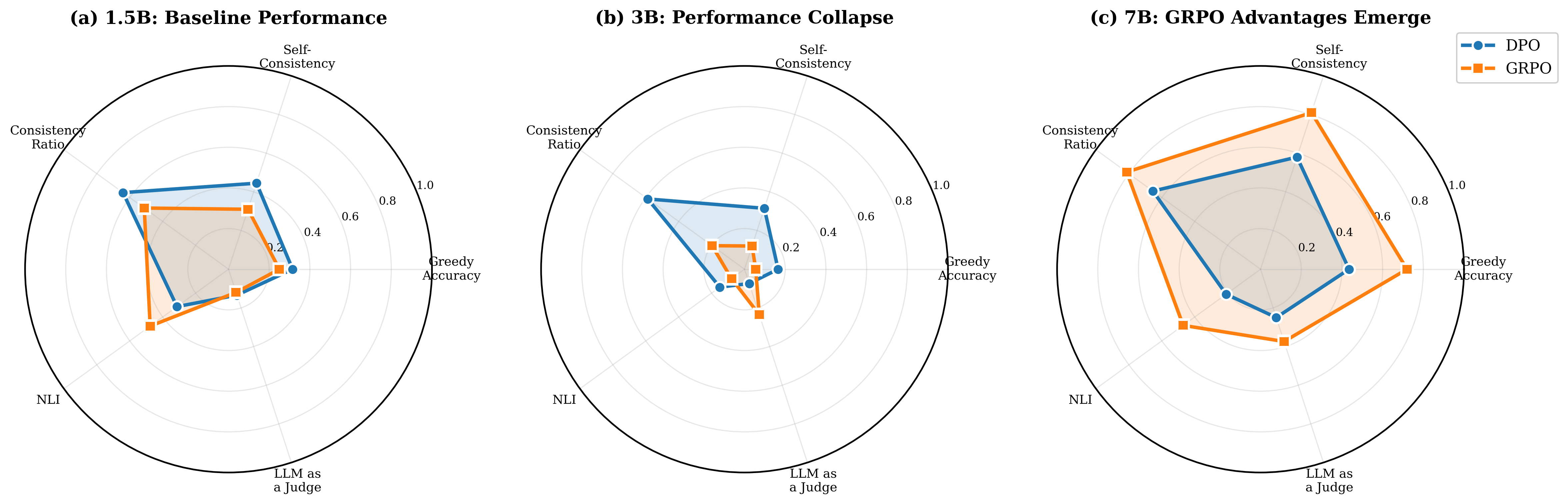}
\caption{Model-scale comparison: (a) 1.5B baseline similarity; (b) 3B performance collapse; (c) 7B GRPO advantages emerge in faithfulness metrics.}
\label{fig:radar}
\end{figure}

\subsection{Faithfulness and Reliability Analysis}

Greedy accuracy, self-consistency, and consistency ratio show strong correlations across all models (Pearson $r>0.85$). This result indicates that the measures capture similar aspects of reliability. Models that answer correctly also tend to do so consistently, producing stable reasoning paths across generations.

Faithfulness behaves differently. DPO’s NLI scores remain almost unchanged as model size increases (from 1.5B: 0.313 to 14B: 0.314), while its LLM-Judge scores rise steadily from 0.134 to 0.576. GRPO, in contrast, improves on both measures: NLI increases from 0.476 to 0.491, and LLM-Judge from 0.120 to 0.748—a fivefold gain. These patterns show that NLI and LLM-Judge capture distinct aspects of faithfulness. NLI tests whether a model’s reasoning logically supports its conclusions, while LLM-Judge reflects the overall quality, coherence, and human alignment of the reasoning process. At the 14B scale, GRPO outperforms DPO by +29.9\% in LLM-Judge and +56.4\% in NLI, suggesting that GRPO produces reasoning that is both more coherent and logically grounded.

Qualitative inspection supports these findings. GRPO learns from multiple reward signals that evaluate structure, correctness, and numerical validity. This setup encourages systematic reasoning rather than memorisation of answer patterns. DPO, on the other hand, trains on preference pairs and focuses on human-like responses. It improves stylistic quality but may not always strengthen internal reasoning consistency. As a result, GRPO tends to generate reasoning that reflects its internal decision process more faithfully, while DPO produces smoother but sometimes less reliable explanations.

These results imply that group-relative optimisation can better align a model’s expressed reasoning with its underlying decision process~\cite{chen-2025,lanham-2023}. However, this advantage appears only at larger scales, where GRPO has enough capacity to stabilise training. The comparison also reveals a practical trade-off: GRPO achieves higher faithfulness but requires careful tuning and greater computational resources, whereas DPO provides steadier improvements that remain accessible on smaller models.

\section{Conclusion}

In this paper, we compared GRPO and DPO for improving faithfulness in chain-of-thought reasoning across Qwen2.5 models ranging from 1.5B to 14B parameters. GRPO achieved the best results at larger scales: the 14B model reached 0.755 greedy accuracy and 0.885 self-consistency, with a +56.4\% gain in NLI faithfulness and +29.9\% in LLM-Judge scores compared to DPO.
Both methods benefited from increased model size, but GRPO displayed higher potential for faithful reasoning alongside less stable behaviour at smaller scales. These results show that GRPO delivers stronger faithfulness at the cost of scale sensitivity and tuning effort, whereas DPO provides steadier, more accessible gains.

Future work should focus on adaptive hyperparameter strategies for smaller models, richer evaluation frameworks for reasoning faithfulness, and more efficient training methods that preserve transparency under resource constraints.

\bibliographystyle{splncs04}
\bibliography{references}







\end{document}